\documentclass[letterpaper, 10 pt, conference]{ieeeconf}  

\IEEEoverridecommandlockouts                              
\usepackage[pdftex]{graphicx}
\usepackage{graphics} 
\usepackage{epsfig} 
\usepackage{mathptmx} 
\usepackage{times} 
\usepackage{amsmath} 
\usepackage{amssymb}  
\usepackage{lipsum}  

\usepackage{caption}

\usepackage{booktabs}
\usepackage[table,xcdraw]{xcolor}

\usepackage{fancyhdr}
\usepackage{url}
\usepackage{verbatim}
\usepackage{xcolor}
\usepackage{listings}

\usepackage[noadjust]{cite}

\usepackage{multicol}

\title{\LARGE \bf
Generating Contextually-Relevant Navigation Instructions\\for Blind and Low Vision People
}

\author{Zain Merchant$^{1}$ \and Abrar Anwar$^{1}$  \and Emily Wang$^{1}$ \and Souti Chattopadhyay$^{1}$ \and  Jesse Thomason$^{1}$ 
\thanks{$^{1}$ Zain Merchant, Abrar Anwar, Emily Wang, Souti Chattopadhyay, and Jesse Thomason are with the Thomas Lord Department of Computer Science, University of Southern California, Los Angeles, CA, USA }
\thanks{Contact: {\tt\small zsmercha@usc.edu}}
}

\begin{document}

\maketitle
\thispagestyle{empty}
\pagestyle{empty}

\begin{abstract}

Navigating unfamiliar environments presents significant challenges for blind and low-vision (BLV) individuals.
In this work, we construct a dataset of images and goals across different scenarios such as searching through kitchens or navigating outdoors.
We then investigate how grounded instruction generation methods can provide contextually-relevant navigational guidance to users in these instances. Through a sighted user study, we demonstrate that large pretrained language models can produce correct and useful instructions perceived as beneficial for BLV users.
We also conduct a survey and interview with 4 BLV users and observe useful insights on preferences for different instructions based on the scenario.

\end{abstract}

\section{INTRODUCTION AND BACKGROUND}

Nearly 253 million people struggle with visual impairment worldwide, where 36 million of these individuals are blind~\cite{ackland2017world}. Dealing with the complexities of daily life poses significant challenges for these individuals, particularly when exploring unfamiliar environments. Traditional aids such as canes and guide dogs are vital in facilitating mobility and independence. However, these tools have limitations in conveying the rich visual information that sighted individuals rely on for navigation and object recognition. 

There has been recent growth in using vision-and-language models as visual assistants that can interactively communicate with a user to provide feedback~\cite{bemyeyes}.
Additionally, after interviews with blind and low vision individuals, prior work has noted a critical issue with the use of guide dogs: the communication from the user to the dog is unimodal. However, there may be questions users want to ask, such as “Is it safe to cross the street?”~\cite{hwang2024towards}. 
If researchers are building robotic guide dogs, the authors posit that any robotic guide dog should handle complex interactions with a user to answer these types of questions.
As the companies and the research community begin to integrate large language models into these robot systems, it is important to understand the role of a language model's contextual understanding capabilities in providing personalized, informative feedback tailored to a user's specific goals and surroundings.
In this work, we focus on navigation assistance and investigate the usefulness and contextual relevance of generated instructions for navigational assistance using large language models (LLMs) and vision-and-language models (VLMs).

Existing work in the field of blind and low vision (BLV) navigation assistance has primarily focused on lower-level navigation tasks such as obstacle avoidance \cite{s19153404, cassinelli2006augmenting, 9187263, Lin_2019_ICCV, bauer2020enhancing}.
Many existing systems gather different kinds of information from the environment, using basic object detection \cite{Leo_2018_ECCV_Workshops} often combined with auditory output \cite{app12052308}  based on templates~\cite{10.1007/978-3-319-16199-0_45} to generate descriptions or simply list objects~\cite{s17112641}. 
While these approaches offer valuable assistance, they often overlook the importance of context and relevance in delivering instructions to BLV users \cite{app12052308, 10.1145/3522757}. 

\begin{figure}
    \centering
    \includegraphics[width=1\linewidth]{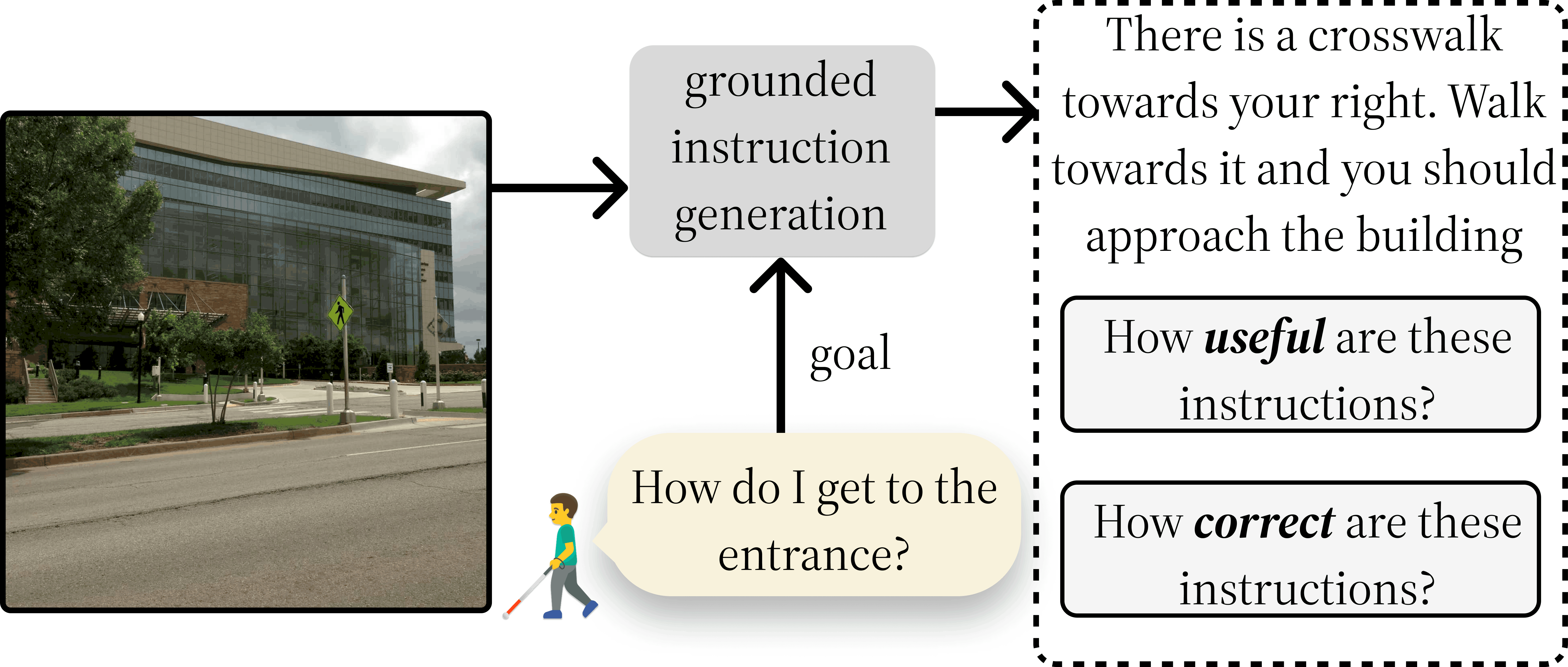}
        \caption{We formulate the problem of providing contextually-relevant navigational instructions to blind and low vision (BLV) people as a grounded instruction generation task, which we then evaluate with sighted and BLV participants in a user study.}
    \label{fig:intro figure}
\end{figure}
\begin{figure*}[t]
    \centering
    \includegraphics[width=.95\textwidth]{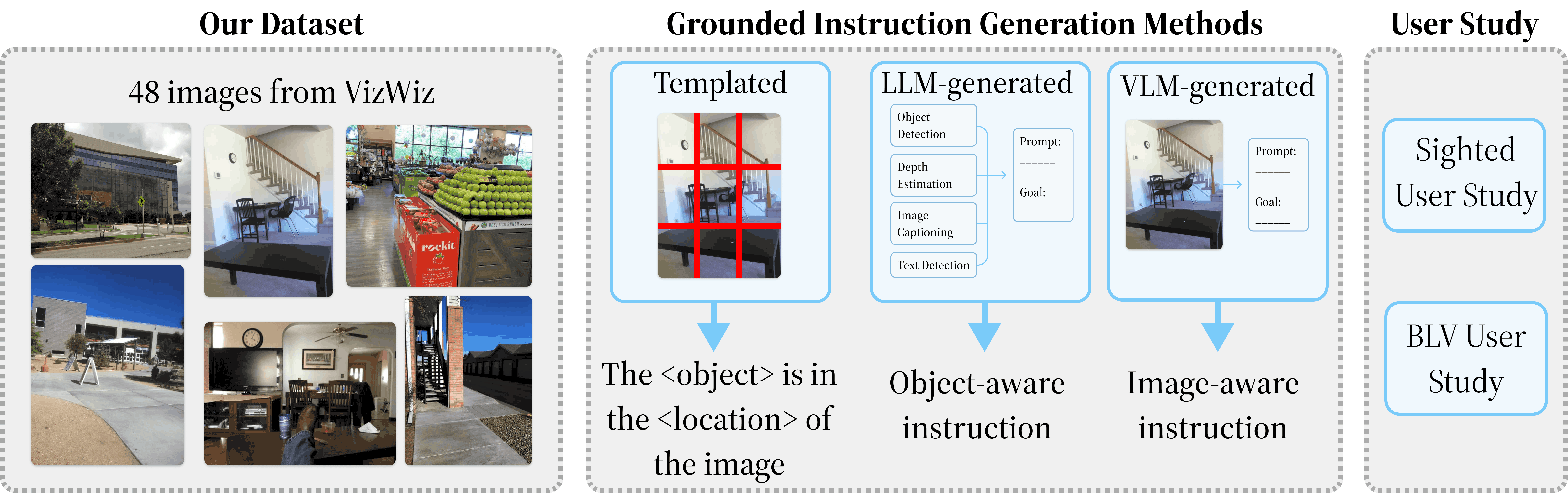}
    \caption{\textit{Left:} We select 48 images from indoor and outdoor environments in VizWiz~\cite{8578478} and annotate them with navigation goals. \textit{Middle:} We design three instruction generation methods, described further in Section~\ref{methods}. \textit{Right:} These generated instructions are then evaluated in a user study with sighted and BLV participants.}
    \label{fig:models}
\end{figure*}

Descriptions are subjective and depend on the user's context \cite{app12052308, kreiss2022context}, and overly generic or unnecessary information is often not preferred by users \cite{10.1145/3027063.3053260, 10.1145/3522757, 10.1145/2661334.2661380}.
Noting these flaws with prior work and considering recent work on the importance of context for BLV participants~\cite{app12052308, 10.1145/3522757}, we propose to augment such systems with LLMs and VLMs and understand whether these methods are able to generate contextually-relevant instructions. 

\section{PROBLEM SETTING}

We want to generate instructions that are \textit{useful} and goal-aware for BLV users, as shown in Figure~\ref{fig:intro figure}. 
We frame this problem as a grounded instruction generation task, where a model $S(w_0, ..., w_N|g, o)$ uses an egocentric image to generate an instruction $w = [w_0, ..., w_N]$  that describes a route for an BLV user to reach a goal $g$ given an image observation $o$. 
The goal $g$ is semantic task context from the user like ``How do I get to the building’s entrance?''
The amount of objects and semantic information that can be conveyed to a user based on an image is often innumerable; however, only a subset of this information is useful to the user to accomplish their goal.
In this work, we are interested in \textit{how the environment and goal impact the usefulness of the generated instruction}. 

VizWiz~\cite{8578478} is a collection of photographs captured by BLV participants and serves as a basis for formulating tasks related to object detection and visual question-answering.

Since VizWiz does not consider navigation instruction generation, we created our own dataset by selecting 48 images from VizWiz that were relevant to navigation from four different environments: offices, kitchens, general indoor, and general outdoor settings. 
We then assigned each image a goal, such as ``Where is the TV remote'', which in navigation tasks, may not be visible in a single frame due to the object's size or placement.

In our setting, we are considering only a single image, but in practical applications, the image alone may not contain the answer to the goal.
For example, in the case of "Where is the TV remote", the remote may not be visible in the image and may require reasoning about potential locations rather than an exact answer.
To capture these kinds of problems, we construct a split such that 20\% of our dataset are \textbf{Hard} examples which require a model to reason more extensively on how to provide instructions for the goal.
The complementary split is referred to as the \textbf{Easy} split.
Each of the images was annotated by sighted volunteers with instructions that would solve the goal. 

\section{Grounded Instruction Generation}
\label{methods}
We compare methods for generating navigation instructions from single images paired with semantic task context.

\textbf{Human instructions.} 
As a point of reference for machine-generated instructions, we sent selected image-goal pairs to four human annotators tasked with generating instructions for BLV users.

\textbf{Template Instructions.}
Past work \cite{10.1007/978-3-319-16199-0_45} extracted the object of interest from the user's goal query. For example, for the sentence ``where is the textbook'', this baseline extracts the object ``textbook'', then detects where the object is using the OWL-ViT open-vocabulary object detector~\cite{minderer2022simple}.
This method then localizes the object into nine predefined areas (top left, center, bottom right, etc.). If a user asks about the location of a microwave, the system can concisely respond: ``The microwave is at the top left.''

\textbf{LLM-generated Instructions.}
Text-only LLMs have good commonsense reasoning abilities given text input; however, unlike VLMs, they cannot take images as input. 
In order to generate grounded navigation instructions, we take a Socratic model \cite{zeng2022socratic} approach, where we provide the outputs of various off-the-shelf models as additional information for the LLM.
We use off-the-shelf object detection~\cite{minderer2022simple}, depth estimation~\cite{DBLP:journals/corr/abs-2201-07436}, image captioning, and optical character recognition.
These inputs are formatted into a prompt with the goal and a few in-context examples.
We use GPT4 \cite{achiam2023gpt} as the LLM and get contextually relevant, grounded instructions as the output.

\textbf{VLM-generated Instructions.}
VLMs are able to take images as input, so rather than a Socratic models approach, we use GPT-4 Vision~\cite{achiam2023gpt} with the image as an input and prompt the model to generate an instruction. 

\section{Study Design}
We conducted two IRB-approved human subject studies with sighted and BLV participants.
\subsection{Sighted User Study}
In addition to the study with BLV participants, we believe that sighted participants can play an important role in validating the correctness of the generated instructions along with gaining an understanding of what they believe is useful.
Sighted participants were asked to take a survey to rate instructions given an image and a goal. 
They rated the instructions in terms of \emph{Correctness} and \emph{Usefulness} on a 1 to 7 Likert scale based on the following definition.
We define \emph{Correctness} as how accurate the instruction is with respect to directions and objects in the image. 
This metric also verifies the instruction generation methods generate accurate instructions.
We define \emph{Usefulness} as how useful the instruction would be to a BLV user to help achieve their goal, considering its relevance and safety.
We recruited eight sighted participants, who each viewed six images across four methods.
We acknowledge the \emph{Usefulness} scores from sighted users are not likely the same as the preferences from a BLV participant; however, since it is difficult to recruit BLV participants at scale, we were interested in also collecting these \emph{Usefulness} scores.
\begin{table}[t]
\centering
\begin{tabular}{lcc}
\toprule
         & \textbf{Correctness}           & \textbf{Usefulness}            \\ \midrule
Templated & $3.85 \pm 1.95$ & $1.96 \pm 1.05$ \\
LLM-based & $4.73 \pm 1.61$ & $4.24 \pm 1.38$ \\
VLM-based    & $4.75 \pm 1.78$ & $4.52 \pm 1.70$ \\
Human    & $5.46 \pm 1.46$ & $4.73 \pm 1.83$ \\ 
\bottomrule
\end{tabular}
\caption{Sighted user ratings for \textbf{Correctness} and \textbf{Usefulness}. Human-annotated instructions are more accurate compared to the other methods, but the LLM- and VLM-generated instructions were rated similarly useful.}
\label{tab:summary_statistics}
\end{table}

\subsection{BLV User Study}
Due to the low incidence of the BLV population, we recruited three blind and one low-vision participant.
Similar to the sighted user study, each image is rated by a BLV participant for \emph{Usefulness}.
We do not collect \emph{Correctness} ratings for the BLV participants since many participants would not be able to compare the instruction to the scene itself due to the level of their visual impairment.
After the survey, we conducted a semi-structured interview.
Survey questions were aimed to elicit their thoughts on navigating different environments, including social spaces, and their thoughts about the kinds of methods they experienced.

\section{Results}

\subsection{Sighted Survey Results}
With our sighted user survey, we find that users find the generated instructions correct and useful, which shows promise for these methods to be tested with BLV users.

\textbf{Generated instructions are similarly useful to human annotated instructions.}
Table~\ref{tab:summary_statistics} shows the aggregated correctness and usefulness scores given by sighted users across methods. 
Users found human-generated instructions more accurate than LLM- and VLM-generated instructions.
In contrast, the difference between usefulness ratings between human, LLM-, and VLM-generated instructions was much smaller.
The difference in usefulness ratings between the VLM and LLM could be explained by harder-to-answer instances benefiting from the input of the entire image, while easy-to-answer instances can be solved more directly with object detectors, as supported by Table~\ref{tab:sighted_easy_useful}.

\begin{table}[t]
\centering
\begin{tabular}{lcc}
\toprule
         & \textbf{Easy} Split           & \textbf{Hard} Split           \\ \midrule
Templated & $2.11 \pm 1.09$ & $1.40 \pm 0.70$ \\
LLM-based    & $4.39 \pm 1.75$ & $5.00 \pm 1.49$ \\
VLM-based & $4.20 \pm 1.30$ & $4.40 \pm 1.71$ \\
Human    & $4.58 \pm 1.80$ & $5.30 \pm 1.95$ \\ \bottomrule
\end{tabular}

\caption{\textbf{Usefulness} scores for the \textbf{Easy} and \textbf{Hard} splits from the 48 image-goal pairs. Interestingly, the \textbf{Easy} split was rated lower than the \textbf{Hard} split.
}
\label{tab:sighted_easy_useful}
\end{table}

\begin{figure}[t]
    \centering
    \includegraphics[width=.9\linewidth]{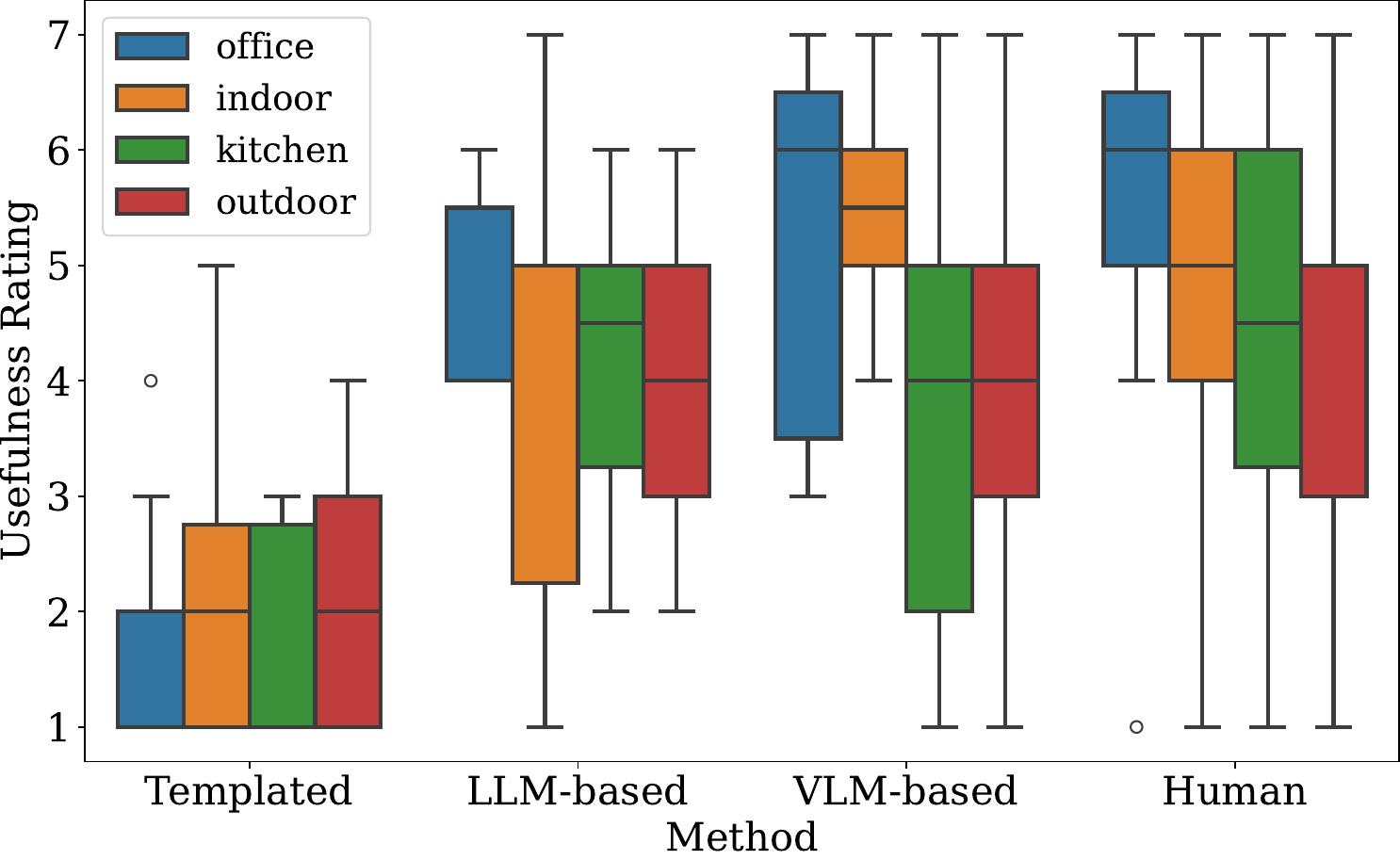}
    \caption{Sighted participant \textbf{Usefulness} ratings over the generated instructions for 48 image-goal pairs across four methods separated by environment. VLM-based instructions had similar ratings across environments to humans. The LLM-based model was rated slightly less useful. }
    \label{fig:box-plot}
\end{figure}

\textbf{Users find different amounts of usefulness of instructions depending on the environment.}
Figure~\ref{fig:box-plot} shows box plots of the usefulness scores of sighted users across different environments. 
We find that the VLM and human-generated instructions have similar usefulness score distributions compared to the LLM-generated and templated instructions.
We also observe that instructions generated for office and general indoor environments are rated more useful than kitchen and outdoor instructions. 
This trend could be because the instruction generation methods have to reason about more complex scenes, or users having different expectations in these scenes.

\subsection{BLV Survey Results}

The sighted user results indicate that the generated instructions are correct and show trends about the role of context.
Though these results provide insights about the generated instructions, we focus on the quantitative and qualitative results from our BLV user study.

\textbf{BLV participants rate methods as less useful compared to sighted participants. }
As shown in Table~\ref{table:blv_vertical}, we find the LLM- and VLM-generated instructions were rated slightly lower to the human. 
Unsurprisingly, the template instructions were consistently not useful. 
Due to the small size of our BLV user study, we will focus primarily on the qualitative semi-structured interview in the next section.


\subsection{BLV Qualitative Interview}

\textbf{Different environments change preferences on the kinds of instructions.} 
Participants indicated that in confined spaces such as a kitchen, they have a preference for less broad instructions, whereas outdoor, open spaces can be more broad (e.g. “walk forward until you reach the corner”). The low-vision participant noted that lighting and audio cues could provide a means for useful guidance.


\textbf{Generative methods rely on visual cues.} 
Participants noted that some responses relied on visual cues such as ``it's near the big sign'' which is not useful. 
Their suggestion was to make the system more specific and to focus on integrating more spatial awareness as those instructions are most useful.

Participants preferred specific directions (e.g. ``take a few steps to your right'', ``10 degrees to the right'') over vague ones (e.g. ``the table is in the center''). 
However, they noted that specificity is not always useful. 
For example, knowing how many objects are on a table might be too much information to be given at once and could be asked as a separate question.

\textbf{What makes a useful navigation assistant?} 
Several participants indicated that an ideal system would tell them where things are laid out in relation to other things that they can reason about. 
For example, ``the sponge is in the bottom right of the basin'' is helpful, but ``the bench is to the right of the sign'' is not. 
In contrast to prior work and systems like Google Maps, participants noted frustration with instructions that stated a precise number of feet to walk, especially since these systems cannot tell the user when they have reached that distance. Instructions like ``walk until you reach the corner'' would resolve this issue.
Thus, leveraging the relationship between the goals with one's surroundings can be helpful.
One participant found the LM-generated instructions to be wordy or condescending, motivating investigations into preferences in \textit{how} these models communicate information.
\begin{table}[t]
\centering
\begin{tabular}{ll}
\toprule
Method & \textbf{Usefulness}\\
\midrule
Templated & $2.00 \pm 1.29$ \\
LLM-based & $3.97 \pm 1.78$ \\
VLM-based    & $4.45 \pm 1.73$ \\
Human    & $4.03 \pm 1.80$ \\
\bottomrule
\end{tabular}
\caption{\textbf{Usefulness} ratings from our BLV participants. The VLM-based instructions were rated as more useful than all of instructions.} 
\label{table:blv_vertical} 
\end{table}

\section{Conclusion, Ethics, and Limitations}
LLM and VLM-based methods for grounded instruction generation show great promise in integrating with assistive technologies. 
However, a significant challenge associated with using these models is their tendency to produce hallucinations or inaccurate generations. 
Poorly generated instructions can lead to confusion and put users in potentially hazardous situations. 

We also recognize the reference instructions written by sighted annotators may not be tailored to how a BLV user may want to be given instructions, as the annotators were not expertly trained to communicate with BLV users.

LLMs and VLMs are also susceptible to biases present in their training data~\cite{srinivasan2021worst}. 
It is important to ensure these technologies are trained on diverse data sets that accurately represent the variety of cultures and environments that may be encountered so that these assistive technologies can serve users in an equitable manner.
By emphasizing the role of context in the generation of instructions for BLV users, we hope our work can initiate a community discussion on how to handle the many possible scenarios a user could experience.








\bibliographystyle{IEEEtran}  
\bibliography{citations}

\clearpage
\clearpage
\section*{APPENDIX}

\label{sec:appendix}
\subsection{LLM Prompt}

I'm going to give you some information about an image. I will then give you a goal. Using the information about the image, I'd like you to give 2-3 sentences telling me how I can achieve my goal and navigating me. This will be used so that a blind person can navigate. 

Rules: 

1. Please give me instructions like left and right and not coordinates. So never use pixel positional information. 

2. You can imagine that the camera is attached to the blind persons head, so the starting point will be the perspective of the camera. 

3. If you don't see the object, note that you don't see it, but succinctly explain where it could be.

4. Responses should be around 2 sentences. 

5. The person using this system might not know much about computers, so use person-centric language. Rather than saying "the detected object", say "I see a" 

6. The objects are printed based on confidence scores. So if an object is printed multiple times, just use the first one's coordinates to help come up with your directions. 

7. Do not discuss how many steps it should take to reach an object, but instead talk about directions. For example, do not say "take 3 steps forward to reach the tv remote" and instead say "the tv remote is in front of you"

Here's an example: The image size is 1500 x 1000

Image caption: a kitchen with a table, chairs, and a refrigerator

Objects:

Detected dining table at location (758, 767) with size 718 x 262 and depth 1 meters

Detected chair at location (529, 739) with size 169 x 229 and depth 1 meters

Detected refrigerator at location (150, 523) with size 295 x 659 and depth 1 meters

Goal: Where is the fridge

Response: The fridge is slightly to your left and forward. Please be careful to not bump into the table and chairs as you move.

The image size is \{\}

Image caption: \{\}

Objects:

Detected \{\} at location \{\} with size \{\} and depth \{\} meters

Text in image: \{\}

Goal: \{\}

\subsection{VLM Prompt:}

I'd like you to give 2-3 sentences telling me how I can achieve my goal and navigating me. This will be used so that a blind person can navigate.

Rules: 

1. Please give me instructions like left and right and not coordinates. So never use pixel positional information.

2. You can imagine that the camera is attached to the blind persons head, so the starting point will be the perspective of the camera. 

3. If you don't see the object, note that you don't see it, but succinctly explain where it could be. 

4. Responses should be around 2 sentences. 

5. The person using this system might not know much about computers, so use person-centric language. Rather than saying "the detected object", say "I see a..."

7. Do not discuss how many steps it should take to reach an object, but instead talk about directions. For example, do not say "take 3 steps forward to reach the tv remote" and instead say "the tv remote is in front of you" Example: The fridge is slightly to your left and forward. Please be careful to not bump into the table and chairs as you move.

\subsection{Semi-structured Interview Questions:}



These are the following questions which we used to guide our semi-structured interviews, which are in several broad categories.

\begin{enumerate}
    \item Streets, homes, grocery stores, etc
    \begin{itemize}
        \item How did you feel about the responses in the context of different environments?
        \begin{itemize}
            \item Streets
            \item Grocery stores
            \item Homes
            
        \end{itemize}
        
        \item What types of concerns did you have when navigating the different environments? [Nudge about safety, completeness, moving components]
        \item Were there moments the responses didn't address your concerns?
        \item Can you share some examples?
        \item What aspects or characteristics of the responses were helpful?

    \end{itemize}
    
    \item Familiar vs unfamiliar environments
    
    \begin{itemize}
        \item What information helps you the most when navigating unfamiliar environments? 
        \item For example, what kinds of objects such as stairs would you find helpful to know when navigating? 
        \item How much detail about these objects would be good to know?
    \end{itemize}
    \item Navigating in social spaces
    \begin{itemize}
        \item What challenges do you face when navigating social spaces like restaurants or events?
        \item What kind of information will help you navigate?
    \end{itemize}
    
    \item Interactions with Others
    \begin{itemize}
        \item How do you prefer others to assist you when navigating in social or public spaces, if at all?
        \item Have you experienced any difficulties or discomfort when seeking assistance from strangers, acquaintances, or online services such as Be My Eyes while navigating?
    \end{itemize}
    
    \item Relevant information in assisted navigation
    
    \begin{itemize}
    
        \item Role of feedback for metric navigation
        \item In what situations do you prefer to navigate independently, and when would you like further assistance?
        \item Consider you are washing dishes, you might have direct information where a system might know where an object is, but sometimes it doesn't. For example, if you want to find a sponge, would it be helpful for the system to tell you about information such as where the dishes or sink is?
    \end{itemize}
    \item Current Technology/Tools
    \begin{itemize}
        \item What technologies or tools do you currently use for navigation?
        \item Are there any specific features or functionalities you wish existing navigation tools had to better assist you? 
        \item Have you encountered any barriers or challenges in using navigation technology or tools?
    \end{itemize}
    \item Emotional Wellbeing
    \begin{itemize}
        \item Have you ever felt overwhelmed or stressed by navigation tasks? What would help you feel less overwhelmed or stressed?
        \item Would reassurance help at all?
        \item What support or resources would you find helpful in managing navigation-related challenges?
    \end{itemize}
    \item Future
    \begin{itemize}
        \item What improvements do you hope to see in navigation assistance for the blind or visually impaired community?
    \end{itemize}
    \item Evaluation of system/generated responses
    \begin{itemize}
        \item You experienced a few different methods we had designed. What would your ideal system that generates instructions look like to you?
        \item Think back to a helpful example. What did you like about those methods and what didn't you like?
        \item Think back to an unhelpful example…
        \item How useful would an ideal system like this be to you?
        \item Do you think systems such as this would help you gain independence? Feels wrong to just ask them this directly.
        \item Were there any other challenges/issues about the methods that you faced?
    \end{itemize}
    \item General Questions
    \begin{itemize}
        \item Do you have any recommendations as to adjustments that would enhance the experience?
        \item Is there anything else you would like to share?
        \item Are there any additional concerns or aspects that you feel as though should be addressed?
    \end{itemize}
\end{enumerate}




\end{document}